\documentclass[conference]{IEEEtran}
\IEEEoverridecommandlockouts
\usepackage[numbers]{natbib}
\usepackage{amsthm}
\usepackage{amsmath,amssymb,amsfonts}

\usepackage[export]{adjustbox}
\usepackage{algorithm,algorithmic}
\usepackage{hyperref}
\usepackage{enumitem}
\usepackage{graphicx}
\usepackage{textcomp}
\usepackage{xcolor}
\usepackage{comment}

\usepackage{subfig}
\usepackage [autostyle, english = american]{csquotes}
\MakeOuterQuote{"}
\def\BibTeX{{\rm B\kern-.05em{\sc i\kern-.025em b}\kern-.08em
    T\kern-.1667em\lower.7ex\hbox{E}\kern-.125emX}}
\newtheorem{definition}{Definition}

\DeclareMathOperator*{\argmin}{arg\,min}
\DeclareMathOperator*{\argmax}{arg\,max}
\begin{document}

\title{Fair Adversarial Gradient Tree Boosting}

\author{
\IEEEauthorblockN{Vincent Grari}
\IEEEauthorblockA{\textit{Sorbonne Universit\'{e}} \\
\textit{LIP6/CNRS} \\
Paris, France \\
vincent.grari@lip6.fr}
\and
\IEEEauthorblockN{Boris Ruf}
\IEEEauthorblockA{\textit{AXA} \\
\textit{REV Research} \\
Paris, France \\
boris.ruf@axa.com}
\and
\IEEEauthorblockN{Sylvain Lamprier}
\IEEEauthorblockA{\textit{Sorbonne Universit\'{e}} \\
\textit{LIP6/CNRS}\\
Paris, France \\
sylvain.lamprier@lip6.fr}
\and 
\IEEEauthorblockN{Marcin Detyniecki}
\IEEEauthorblockA{\textit{AXA} \\
\textit{REV Research} \\
Paris, France \\
marcin.detyniecki@axa.com}
}


\maketitle

\begin{abstract}
Fair classification has become an important topic in machine learning research. While most bias mitigation strategies focus on neural networks, we noticed a lack of work on fair classifiers based on decision trees even though they have proven very efficient. In an up-to-date comparison of state-of-the-art classification algorithms in tabular data, tree boosting outperforms deep learning~\citep{10.1016/j.eswa.2017.04.003}. For this reason, we have developed a novel approach of adversarial gradient tree boosting. The objective of the algorithm is to predict the output $Y$ with gradient tree boosting while minimizing the ability of an adversarial neural network to predict the sensitive attribute $S$. The approach incorporates at each iteration the gradient of the neural network directly in the gradient tree boosting. We empirically assess our approach on 4 popular data sets and compare against state-of-the-art algorithms. The results show that our algorithm achieves a higher accuracy while obtaining the same level of fairness, as measured using a set of different common fairness definitions.
\end{abstract}


\section{Introduction}

Machine learning models are increasingly used in decision making processes. In many fields of application, they generally deliver superior performance compared with conventional, deterministic algorithms. However, those models are mostly black boxes which are hard, if not impossible, to interpret. Since many applications of machine learning models have far-reaching consequences on people (credit approval, recidivism score etc.), there is growing concern about their potential to reproduce discrimination against a particular group of people based on sensitive characteristics such as gender, race, religion, or other. In particular,  algorithms trained on biased data are prone to learn, perpetuate or even reinforce these biases~\citep{NIPS2016_6228}. In recent years, many incidents of this nature have been documented. For example, an algorithmic model used to generate predictions of criminal recidivism in the United States (COMPAS) discriminated against black defendants~\citep{angwin2016machine}. Also, discrimination based on gender and race could be demonstrated for targeted and automated online advertising on employment opportunities~\citep{Lambrecht}. In this context, the EU introduced the General Data Protection Regulation (GDPR) in May 2018. This legislation represents one of the most important changes in the regulation of data privacy in more than 20 years. It strictly regulates the collection and use of sensitive personal data. With the aim of obtaining non-discriminatory algorithms, it rules in Article 9(1): "Processing of personal data revealing racial or ethnic origin, political opinions, religious or philosophical beliefs, or trade union membership, and the processing of genetic data, biometric data for the purpose of uniquely identifying a natural person, data concerning health or data concerning a natural person's sex life or sexual orientation shall be prohibited."~\citep{citeulike:14071352}. One fairness method often used in practice today is to remove protected attributes from the data set. This concept is known as "fairness through unawareness"~\citep{Pedreshi2008}. While this approach may prove viable when using conventional, deterministic algorithms with a manageable quantity of data, it is insufficient for machine learning algorithms trained on "big data". Here, complex correlations in the data may provide unexpected links to sensitive information. This way, presumably non-sensitive attributes can serve as substitutes or proxies for protected attributes. 

For this reason, next to optimizing the performance of a machine learning model, the new challenge for data scientists is to determine whether the model output predictions are discriminatory, and how they can mitigate such unwanted bias as much as possible.

Many bias mitigation strategies for machine learning have been proposed in recent years, however, most of them focus on neural networks. Ensemble methods combining several decision tree classifiers have proven very efficient for various applications. Therefore, in practice for tabular data sets, actuaries and data scientists prefer the use of gradient tree boosting over neural networks due to its generally higher accuracy rates. Our field of interest is the development of fair classifiers based on decision trees. In this paper, we propose a novel approach to combine the strength of gradient tree boosting with an adversarial fairness constraint.
The contributions of this paper are threefold:
\begin{itemize}
    \item To the best of our knowledge, we propose the first adversarial learning method for generic classifiers, including decision trees; 
    \item We apply adversarial learning for fair classification on decisions trees;
    \item We empirically compare our proposal and its variants with several state-of-the-art approaches, for two different fairness metrics. Experiments show the great performance of our approach. 
\end{itemize}

The remainder of this paper proceeds as follows. First, section \ref{sec:definitions_of_fairness} presents our notation and introduces common definitions of fairness which will serve as metrics to measure the performance of our approach. Then, section \ref{sec:rw}  reviews papers related with our work. Section \ref{sec:gradient_tree_boosting} briefly recaps the principle of classical gradient tree boosting. 
Next, 
section \ref{sec:Fair Adversarial Gradient Tree Boosting} outlines a novel algorithm which combines gradient tree boosting with adversarial debiasing. Finally, section \ref{sec:empirical_results} 
presents experimental results of our approach.


\section{Fair Machine Learning}

\subsection{Definitions of Fairness}
\label{sec:definitions_of_fairness}

Throughout this document, we consider a classical supervised classification problem training with $n$ examples 
${(x_{i},s_{i},y_{i})}_{i=1}^{n}$, where $x_{i} \in \mathbf{R}^{p}$ is the feature vector with $p$ predictors of the $i$-th example, $s_i$ is its 
binary sensitive attribute and 
$y_{i}$ its binary label. 

In order to achieve fairness, it is essential to establish a clear understanding of its formal definition. In the following we outline the most popular definitions used in recent research. First, there is information sanitization which limits the data that is used for training the classifier. Then, there is individual fairness, which binds at the individual level and suggests that fairness means that similar individuals should be treated similarly. Finally, there is statistical or group fairness. This kind of fairness partitions the world into groups defined by one or several high level sensitive attributes. It requires that a specific relevant statistic about the classifier is equal across those groups. In the following, we focus on this family of fairness measures and explain the most popular definitions of this type used in recent research. 

\subsubsection{Demographic Parity}
\label{sec:demographic_parity}
Based on this definition, a classifier is considered fair if the prediction $\widehat{Y}$ from features $X$ is independent from the protected attribute $S$~\citep{Dwork2011}. The underlying idea is that each demographic group has the same chance for a positive outcome.

\begin{definition}
$P(\widehat{Y}=1|S=0)=P(\widehat{Y}=1|S=1)$
\end{definition}



There are multiple ways to assess this objective.
The p-rule assessment ensures the ratio of the positive rate for the unprivileged group is no less than  a fixed threshold $\frac{p}{100}$.
The classifier is considered as totally fair when this ratio satisfies a 100\%-rule. Conversely, a 0\%-rule indicates a completely unfair model.
\begin{equation}
\emph{P-rule: }
\min(\frac{P(\widehat{Y}=1|S=1)}{P(\widehat{Y}=1|S=0)},\frac{P(\widehat{Y}=1|S=0)}{P(\widehat{Y}=1|S=1)})
\end{equation}
The second metric available for demographic parity is the disparate impact (DI) assessment~\citep{Feldman2014}. It considers the absolute difference of outcome distributions for sub populations with different sensitive attribute values. The smaller the difference, the fairer the model.
\begin{equation}
\emph{DI: }
|P(\widehat{Y}=1|S=1)-P(\widehat{Y}=1|S=0)|
\end{equation}
\subsubsection{Equalized Odds}
\label{sec:equalized_odds}
An algorithm is considered fair if across both demographics $S=0$ and $S=1$, for the outcome $Y=1$ the predictor $\widehat{Y}$ has equal \textit{true} positive rates, and for $Y=0$ the predictor $\widehat{Y}$ has equal \textit{false} positive rates~\citep{Hardt2016}. This constraint enforces that accuracy is equally high in all demographics since the rate of positive and negative classification is equal across the groups. The notion of fairness here is that chances of being correctly or incorrectly classified positive should be equal for every group.
\begin{definition}
$P(\widehat{Y}=1|S=0,Y=y)=P(\widehat{Y}=1|S=1,Y=y), \forall y\in\{0,1\}$
\end{definition}




A metric to assess this objective is to measure the disparate mistreatment (DM)~\citep{Zafar2017}. It computes the absolute difference between the false positive rate (FPR) and the false negative rate (FNR) for both demographics.
{\small
\begin{eqnarray}
D_{FPR}:
|P(\widehat{Y}=1|Y=0,S=1)-P(\widehat{Y}=1|Y=0,S=0)|\\
D_{FNR}:
|P(\widehat{Y}=0|Y=1,S=1)-P(\widehat{Y}=0|Y=1,S=0)|
\end{eqnarray}
}
The closer the values of $D_{FPR}$ and $D_{FNR}$ to 0, the lower the degree of disparate mistreatment of the classifier.

%
%
%

\subsection{Related Work} 
\label{sec:rw}
Recently, research in fair machine learning has prospered, and considerable progress was made when it comes to quantifying and mitigating undesired bias. For the mitigation strategies, 3 distinct approaches exist.

Algorithms which belong to the "pre-processing" family ensure that the input data is fair. This can be achieved by suppressing the sensitive attributes, by changing class labels of the data set, and by reweighting or resampling the data~\citep{Kamiran2012,Bellamy2018,Calmon2017}.

The second type of mitigation strategies comprises the "in-processing" algorithms. Here, undesired bias is directly mitigated during the training phase. A straightforward approach to achieve this goal is to integrate a fairness penalty directly in the loss function. One such algorithm integrates a decision boundary covariance constraint for logistic regression or linear SVM~\citep{Zafar2015}. In another approach, a meta algorithm takes the fairness metric as part of the input and returns a new classifier optimized towards that fairness metric~\citep{DBLP:journals/corr/abs-1806-06055}. Furthermore, the emergence of generative adversarial networks (GANs) provided the required underpinning for fair classification using adversarial debiasing~\citep{NIPS2014_5423}. In this field, a neural network classifier is trained to predict the label $Y$, while simultaneously minimizing the ability of an adversarial neural network to predict the sensitive attribute $S$~\citep{Zhang2018,Wadsworth2018,Louppe2016}.

The final group of mitigation algorithms follows a "post-processing" approach. In this case, only the output of a trained classifier is modified. A Bayes optimal equalized odds predictor can be used to change output labels with respect to an equalized odds objective~\citep{Hardt2016}. A different paper presents a weighted estimator for demographic disparity which uses soft classification based on proxy model outputs~\citep{Chen2019}. The advantage of post-processing algorithms is that fair classifiers are derived without the necessity of retraining the original model which may be time consuming or difficult to implement in production environments. However, this approach may have a negative effect on accuracy or could compromise any generalization acquired by the original classifier~\citep{Donini2017}.

\begin{algorithm}[h]
   \caption{Classical Gradient Boosting}
   \label{alg:classical_gradient_boosting}
   \begin{enumerate}[label={},leftmargin=0.2cm]
   \item {\bfseries Input:} Training set ${(x_{i},s_{i},y_{i})}_{i=1}^{n},$ a number of iterations $M$, a differentiable loss function ${\cal L}(y,F(x))$
   \item {\bfseries Initialize:} Calculate the constant value:
   \begin{flalign*}
     F_{0}(x)=\operatorname*{arg\,min}_\gamma\sum_{i=1}^{n}\mathcal{L}(y_{i},\gamma) 
   \end{flalign*}
   \item {\bfseries for} $m=1$ {\bfseries to} $M-1$ {\bfseries do}
   \begin{enumerate}[label=(\alph*),leftmargin=0.9cm,labelsep=0.3cm]
   \item Calculate the pseudo residuals:
     \begin{flalign*}
     \begin{split}
     r_{im}=-\left[{\frac {\partial \mathcal{L}(y_{i},F(x_{i}))}{\partial F(x_{i})}}\right]_{F(x)=F_{m-1}(x)}\\
     \quad {\mbox{for }}i=1,\ldots,n
     \end{split}
     \end{flalign*}
     \item Fit a classifier $h_{m}(x)$ to pseudo residuals using the training set ${(x_{i},r_{im})\}_{i=1}^{n}}$
     \item Compute multiplier $\gamma_{m}$ by solving the following one-dimensional optimization problem:
     \begin{equation*}
     \gamma_m =  \arg\min_{\gamma}\sum_{i=1}^n \mathcal{L}\left(y_i, F_{m-1}(x_i) + \gamma*h_m(x_i)\right)
     \end{equation*}
     \item Update the  model:
     \begin{equation*}
     F_{m}(x_i) = F_{m-1}(x_i) + \gamma_m * h_m(x_i)
     \end{equation*}
   \end{enumerate}
   \item \bfseries{end for}
   \end{enumerate}

\end{algorithm}

\section{Gradient Tree Boosting}
\label{sec:gradient_tree_boosting}
In order to establish the basis for our approach and also to introduce our notation, we first summarize the principle of classical gradient  tree  boosting. The "Gradient Boosting Machine" (GBM) constitutes a prediction model for regression and classification problems based on an ensemble technique where multiple weak learners are combined to produce a strong learner~\citep{Friedman2001}. Often, such weak learners are decision trees, generally of the type Classification And Regression Tree (CART). In this case, the algorithm is called gradient tree boosting (GTB). The weak learners are built sequentially. Eventually, a strong classifier is obtained  as a weighted sum of the weak learners. The classical gradient descent algorithm is used to optimize the model by any differentiable loss function.

The objective of the GBM is to find a good estimate of the function $f$ which approximately minimizes the empirical loss function:
\begin{flalign}
    \min_F{ \sum_{i=1}^n \mathcal{L}(y_{i},F(x_{i}))} 
\end{flalign}

where the loss function $\mathcal{L}(y_{i}, F(x_{i}))$ measures the $i$-th prediction compared to the true label. In the classical version of the GBM, the prediction corresponding to a feature vector $x$ is given by an additive model of the form

\begin{flalign}
    F_M(x_i) = \sum_{m=0}^M \gamma_m h_m(x_i)
\end{flalign}
where $M$ is the total number of iterations, and $h_m(x_i)$ corresponds to a weak learner at step $m$ in the form of a greedy CART prediction. 

The main steps for fitting the model are shown as pseudo code in Algorithm~\ref{alg:classical_gradient_boosting}. The method exploits the fact that the residual corresponds to the negative gradient of the loss function. Thus, we calculate at each step $m$ the so-called "pseudo residuals":

\begin{flalign}
r_{im}=-\left[{\frac {\partial \mathcal{L}(y_{i},F(x_{i}))}{\partial F(x_{i})}}\right]_{F(x)=F_{m-1}(x)}\quad {\mbox{for }}i=1,\ldots,n
\end{flalign}

In order to update the model, we fit a new weak learner $h_m(x)$ to those pseudo residuals and add it to the current model. This step is repeated until the algorithm converges.

\begin{figure*}[h]
\begin{center}
    \includegraphics[width=17cm]{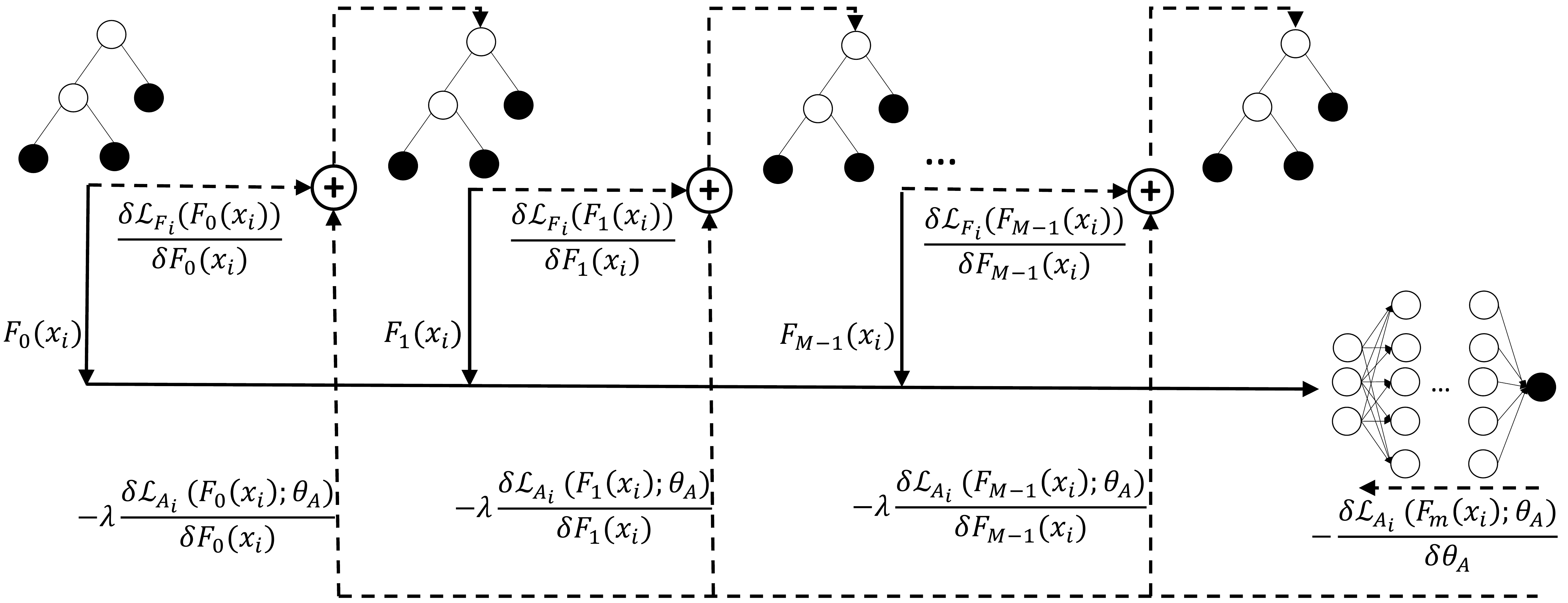}
\caption{The architecture of the Fair Adversarial Gradient Tree Boosting (FAGTB). 4 steps are depicted, each one corresponding to a tree $h$ that is added to the global classifier $F$. The neural network on the right is the adversary that tries to predict the sensitive attributes from the outputs of the classifier. Solid lines represents forward operations, while dashed ones represent gradient propagation. At each step $m$, gradients from the prediction loss and the adversary loss are summed to form the target for the next decision tree $h_{m+1}$.   
}
\label{fig:schema}
\end{center}
\end{figure*}

\section{Fair Adversarial Gradient Tree Boosting (FAGTB)}
\label{sec:Fair Adversarial Gradient Tree Boosting}

Our aim is to learn a classifier that is both effective for predicting true labels and fair, in the sense that it cares about metrics defined in section \ref{sec:definitions_of_fairness} for demographic parity or equalized odds. The idea is to leverage the great performance of GTB for classification, while adapting it for fair machine learning via adversarial learning. 


\subsection{Min-Max formulation}

While most state of the art algorithms focus on the independence of the predicted probability predictions

The GTB processes sequentially by gradient iteration (see Section \ref{sec:gradient_tree_boosting}). This architecture allows us to apply for fair classification with decision tree algorithms the concept of  adversarial learning, which corresponds to a two-player game with two contradictory components, such as in generative adversarial  networks (GAN) \cite{Goodfellow2014}.  
In the vein of ~\citep{Zhang2018,Louppe2016,Wadsworth2018} for fair classification, we consider a predictor function $F$, that outputs the probability of an input vector $X$ for being labelled $Y=1$, 
and an adversarial model $A$ which tries to predict the sensitive attribute $S$ from the output of $F$. Depending on the accuracy rate of the adversarial algorithm, we penalize the gradient of the GTB at each iteration. The goal is to obtain a classifier $F$ whose outputs 
do not allow the adversarial function 
to reconstruct the value of the sensitive attribute. If this objective is achieved, the data bias in favor of some demographics disappeared from the output prediction.


The predictor and the adversarial classifiers are optimized simultaneously in a min-max game defined as: 
\begin{eqnarray}
    \argmin_F\max_{\theta_{A}} \sum\limits_{i=1}^n \mathcal{L}_{F_i}(F(x_i))-\lambda \sum\limits_{i=1}^n \mathcal{L}_{A_i}(F(x_i) ; \theta_{A})
    \label{valuefunction}
\end{eqnarray}
where $\mathcal{L}_{F_i}$ and $\mathcal{L}_{A_i}$ are respectively the predictor and the adversary loss for the training sample $i$ given $F(x_i) \in \mathbb{R}$, which refers to the output of the GTB predictor for input $x_i$. The hyperparameter $\lambda$ controls the impact of the adversarial loss.

The targeted classifier outputs  the label $\widehat{Y}$ which maximizes the posterior $P(\widehat{Y}|X)$. Thus, for a  given sample $x_i$, we get:
\begin{equation}
\hat{y}_i=\argmax_{y \in \{0;1\}} p_{F}(Y=y|X=x_i)
\end{equation}
where $p_{F}(Y=1|X=x_i)=\sigma(F(x_i))$, with $\sigma$ denoting the sigmoid function. Therefore,  
$\mathcal{L}_{F_i}$ is defined as the negative log-likelihood of the predictor for the training sample $i$: 
\begin{eqnarray}
    \begin{aligned}
    \mathcal{L}_{F_i}(F(x_i)) =& -\log p_{F}(Y=y_i|X=x_i) \\ =&- \mathbf{1}_{y_i=1} \log(\sigma(F(x_i))) \\&   - \mathbf{1}_{y_i=0} \log(1-\sigma(F(x_i)))
    \end{aligned}
\end{eqnarray}
where $\mathbf{1}_{cond}$ equals 1 if $cond$ is true, 0 otherwise. 

The adversary $A$ corresponds to a neural network  with parameters $\theta_A$, which takes as input the sigmoid of the predictor's output for any sample $i$  (i.e., $P_{F}(Y=1|X=x_i)$), and outputs the probability $P_{F,\theta_{A}}$ for the sensitive to equal 1:  
\begin{itemize}
\item For the demographic parity task, $P_{F}(Y=1|X=x_i)$ is the only input given to the adversary for the prediction of the sensitive attribute $s_i$. In that case, the network $A$ outputs the conditional probability $P_{F,\theta_{A}}(S=1|V=v_i)=A(v_i)$, with $V=(\sigma(F(X)))$. 
\item For the equalized odds task, the label $y_i$ is concatenated to $P_{F}(Y=1|X=x_i)$ to form the input vector of the adversary $v_i=(\sigma(F(x_i)),y_i)$, so that the function $A$ could be able to output different conditional probabilities $P_{F,\theta_{A}}(S=1|V=v_i)$ depending on the label $y_i$ of $i$.  
\end{itemize}
The adversary loss is then defined for any training sample $i$ as:
\begin{eqnarray}
    \begin{aligned}
    \mathcal{L}_{A_i}(F(x_i) ; \theta_{A}) =& -  \mathbf{1}_{s_i=1} \log(\sigma(A(v_i))) \\&   - \mathbf{1}_{s_i=0} \log(1-\sigma(A(v_i)))
    \end{aligned}
\end{eqnarray}
with $v_i$ defined according to the task as detailed above. 

Note that, for the case of demographic parity, if there exists $(F^*,\theta_{A}^*)$ such that $\theta_{A}^*=\argmax_{\theta_{A}} P_{F^*,\theta_{A}}(S|V)$ on the training set, $P_{F^*,\theta_{A}^*}(S|V)=\widehat{P}(S)$ and $P_{F^*}(Y|X)=\widehat{P}(Y|X)$, with $\widehat{P}(S)$ and $\widehat{P}(Y|X)$ the corresponding distributions on the training set, $(F^*,\theta_{A}^*)$ is a global optimum of our min-max problem eq. (\ref{valuefunction}). In that case, we have both a perfect classifier in training, and a completely fair model since 
the best possible adversary is not able to predict $S$ more accurately  than the estimated prior distribution. 
Similar observations can easily be made for the equalized odds task (by replacing $\widehat{P}(S)$ by $\widehat{P}(S|Y)$ and using the corresponding definition of $V$ in the previous assertion). While such a perfect setting does not always exists in the data, it shows that the model is able to identify a solution when it reaches one. 
If a perfect solution does not exists in the data, the optimum of our min-max problem is a trade-off between prediction accuracy and fairness, controlled by the hyperparameter $\lambda$. 

\subsection{Learning} 

The learning process is outlined as pseudo code in Algorithm~\ref{alg:fair_adversarial_gradient_tree_boosting}.
The algorithm first initializes the classifier $F_0$ with constant values for all inputs, as done for the classical GBT. Additionally, it initializes the parameters $\theta_A$ of the adversarial neural network $A$ (a Xavier initialization is used in our experiments’). Then, at each iteration $m$, beyond calculating the pseudo residuals $r_{im}$ for any training sample $i$ w.r.t. the targeted prediction loss $\mathcal{L}_{F_i}$, it computes pseudo residuals $t_{im}$ for the adversarial loss $\mathcal{L}_{A_i}$ too. Both residuals are combined in $u_{im}= r_{im}-\lambda*t_{im}$, where $\lambda$ controls the impact of the adversarial network. The algorithm then fits a new weak regressor $h_m$ (a decision tree in our work) to residuals using the training set $\{(x_{i},u_{im})\}_{i=1}^{n}$. This pseudo-residuals regressor is supposed to correct  both prediction and adversarial biases of the old classifier $F_{m-1}$. It is added to it after a line search step, which determines the best $\gamma_m$ weight to assign to $h_m$ in the new classifier $F_m$. Finally, the adversarial has to adapt its weights according to new outputs (i.e., using the training set $\{(F_{m}(x_i),s_{i})\}_{i=1}^{n}$). This is done by gradient backpropagation. A schematic representation of our approach can be found in Figure~\ref{fig:schema}. 





\begin{algorithm}[!h]
   \caption{Fair Adversarial Gradient Tree Boosting}
   \label{alg:fair_adversarial_gradient_tree_boosting}
   \begin{enumerate}[label={},leftmargin=0.2cm]
   \item {\bfseries Input:} training set ${(x_{i},s_{i},y_{i})}_{i=1}^{n},$ a number of iterations M, an adversarial learning rate $\alpha$, a differentiable loss function 
${\cal L}_{F}$ for the output classifier and ${\cal L}_{A}$ for the adversarial classifier.\\
   \item {\bfseries Initialize:} Calculate the constant value:
    \begin{flalign*}
    F_{0}(x)=\operatorname*{arg\,min}_\gamma\sum_{i=1}^{n}{\cal L}_{F_i}(\gamma) 
    \end{flalign*}
    Initialize parameters $\theta_{A}$ of the neural network $A(x)$\\
   \item {\bfseries for} $m=1$ {\bfseries to} $M-1$ {\bfseries do}
   \begin{enumerate}[label=(\alph*),leftmargin=0.9cm,labelsep=0.3cm]
     \item
     Calculate the pseudo residuals:
    \begin{flalign*}
    \begin{split}
    r_{im}=-\left[{\frac {\partial {\cal L}_{F_i}(F(x_{i}))}{\partial F(x_{i})}}\right]_{F(x)=F_{m-1}(x)}\\
    \quad {\mbox{for }}i=1,\ldots ,n
    \end{split}
    \end{flalign*}
    \item
    Calculate the pseudo residuals of the adversarial from the input $F_{m-1}(x_{i})$: 
    \begin{flalign*}
    \begin{split}
    t_{im}=-\left[{\frac {\partial {\cal L}_{A_i}(F(x_{i};\theta_{A}))}{\partial F(x_{i})}}\right]_{F(x)=F_{m-1}(x)}\\
    \quad {\mbox{for }}i=1,\ldots,n
    \end{split}
    \end{flalign*}
    \item
    Calculate the training loss derivative:
    \begin{flalign*}
    u_{im}= r_{im}-\lambda*t_{im}
    \end{flalign*}
    \item 
    Fit a classifier $h_{m}(x)$ to pseudo residuals using the training set $\{(x_{i},u_{im})\}_{i=1}^{n}$
    \item
    Compute multiplier $\gamma_{m}$ by solving the following one-dimensional optimization problem:
      \begin{equation*}
      \begin{split}
        \gamma_m =\arg\min_{\gamma}\sum_{i=1}^n {\cal L}_{F_i}\left(F_{m-1}(x_i) +  \gamma*h_m(x_i)\right) \\
         -\lambda*{\cal L}_{A_i}(F_{m-1}(x_i) +  \gamma*h_m(x_i);\theta_{A}).
      \end{split}
      \end{equation*}
    \item
    Update the learning model:
    \begin{equation*}
    F_{m}(x_i) = F_{m-1}(x_i) + \gamma_m * h_m(x_i)
    \end{equation*}
    \item
    Fit the adversarial $A$ to the using the new outputs (i.e., using the training set  $\{(F_{m}(x_i),s_{i})\}_{i=1}^{n}$)
    \begin{equation*}
    \theta_{A} := \theta_{A} - \alpha * {\frac {\partial {\cal L}_{A_i}(F_{m}(x_i);\theta_{A})}{\partial\theta_{A}}}
    \end{equation*} 
    \end{enumerate}

   \item {\bfseries end do}
   \end{enumerate}

\end{algorithm}
    

 \begin{figure*}[h]
  \centering
  \includegraphics[scale=0.38,valign=t]{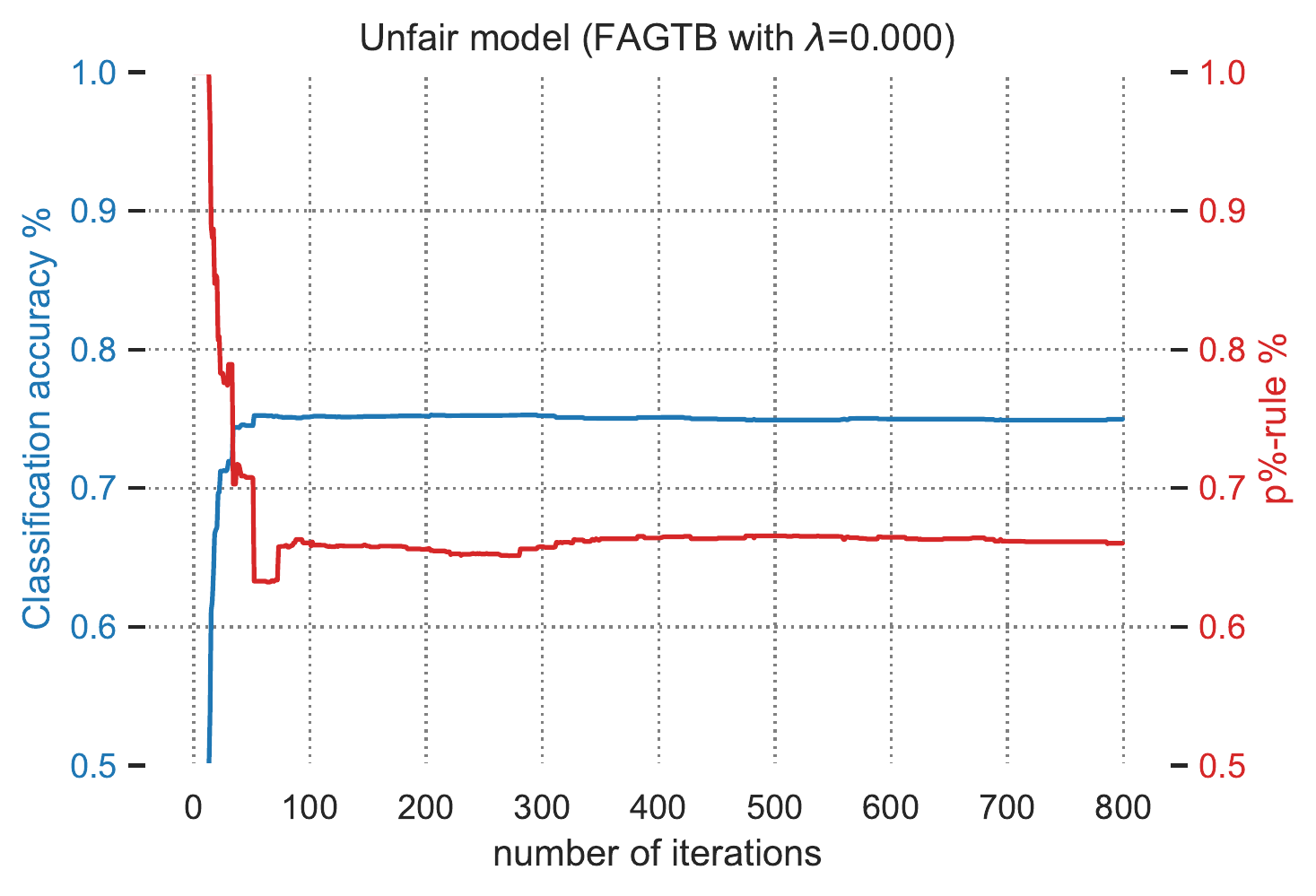}
  \includegraphics[scale=0.38,valign=t]{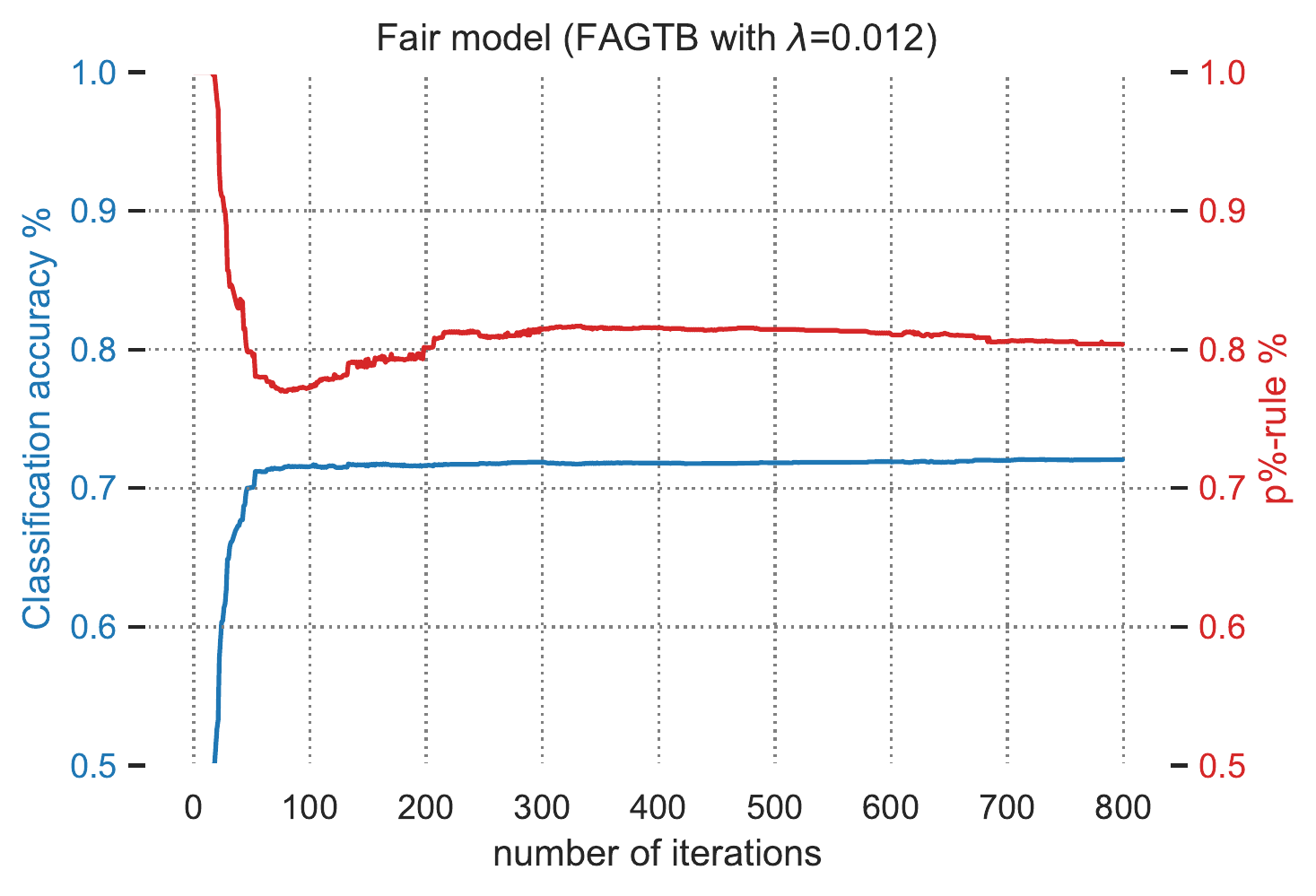}
  \includegraphics[scale=0.38,valign=t]{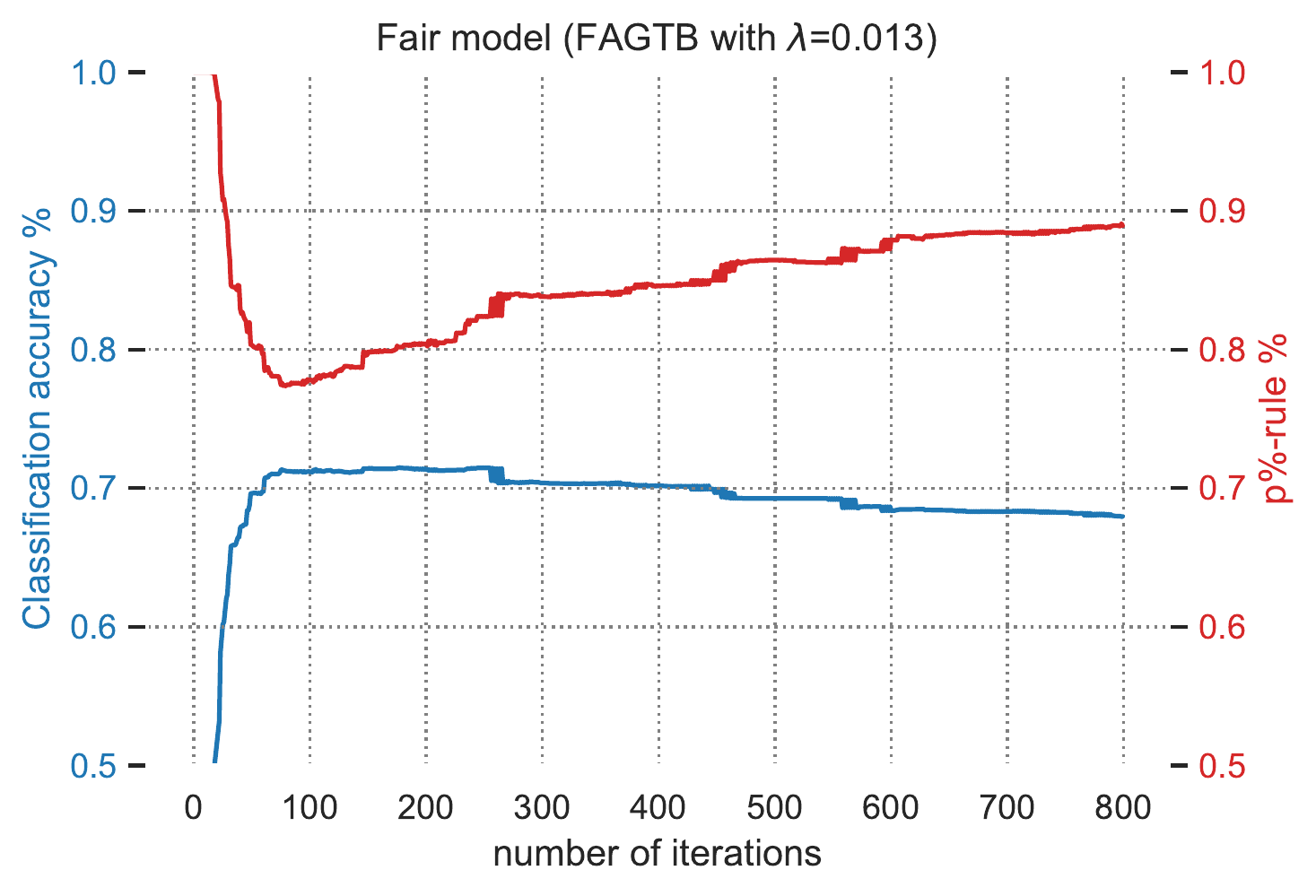}
  \includegraphics[scale=0.38,valign=t]{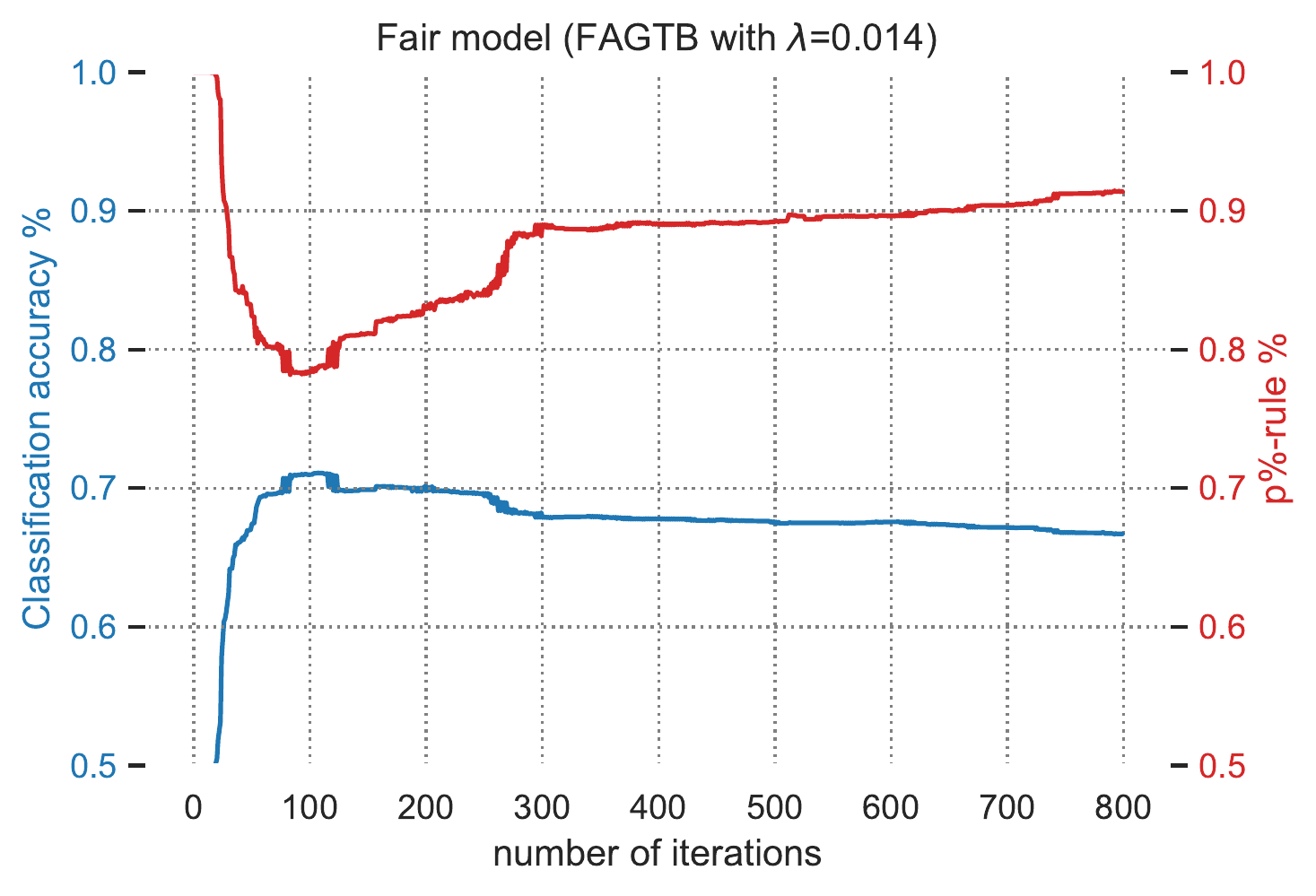}
  \includegraphics[scale=0.38,valign=t]{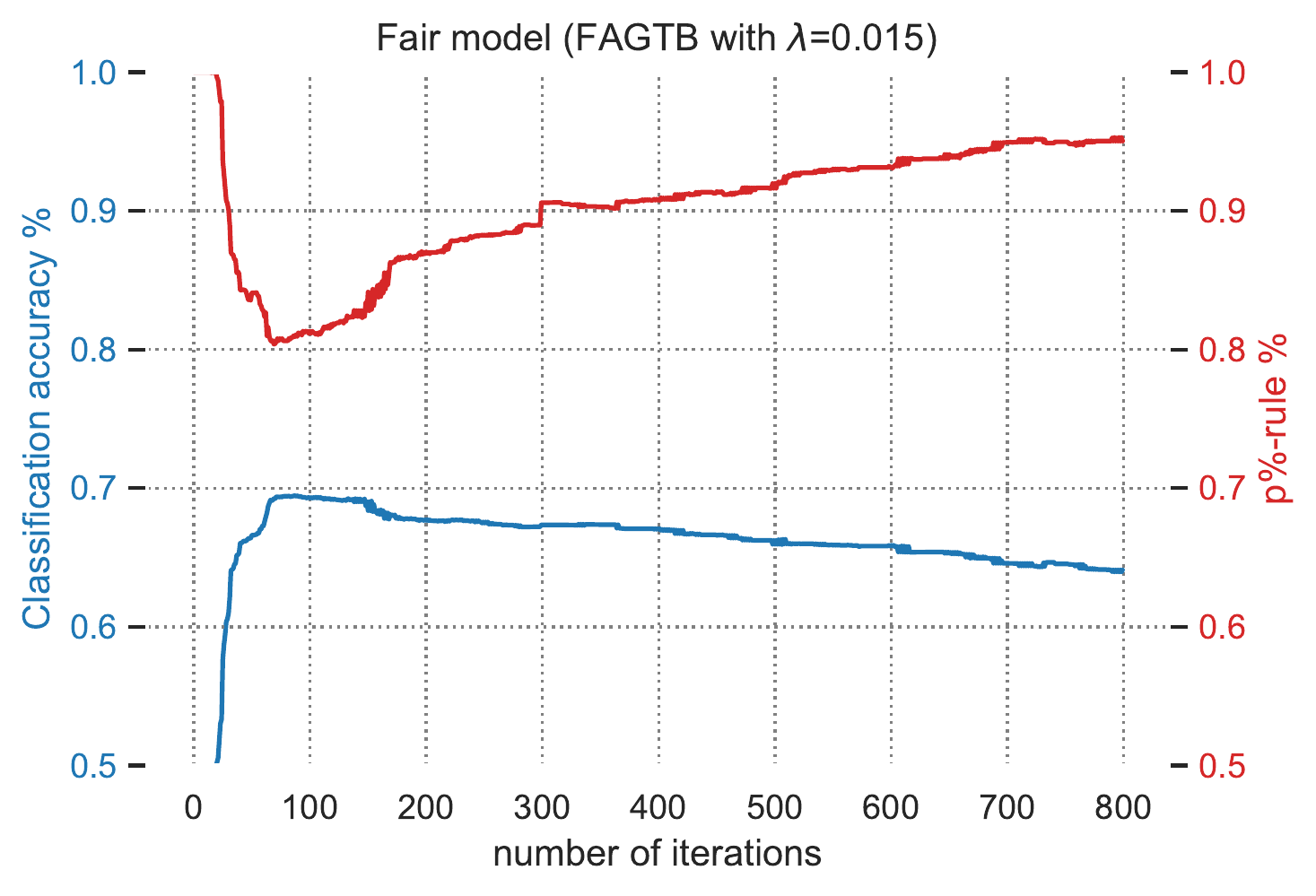}
  \includegraphics[scale=0.38,valign=t]{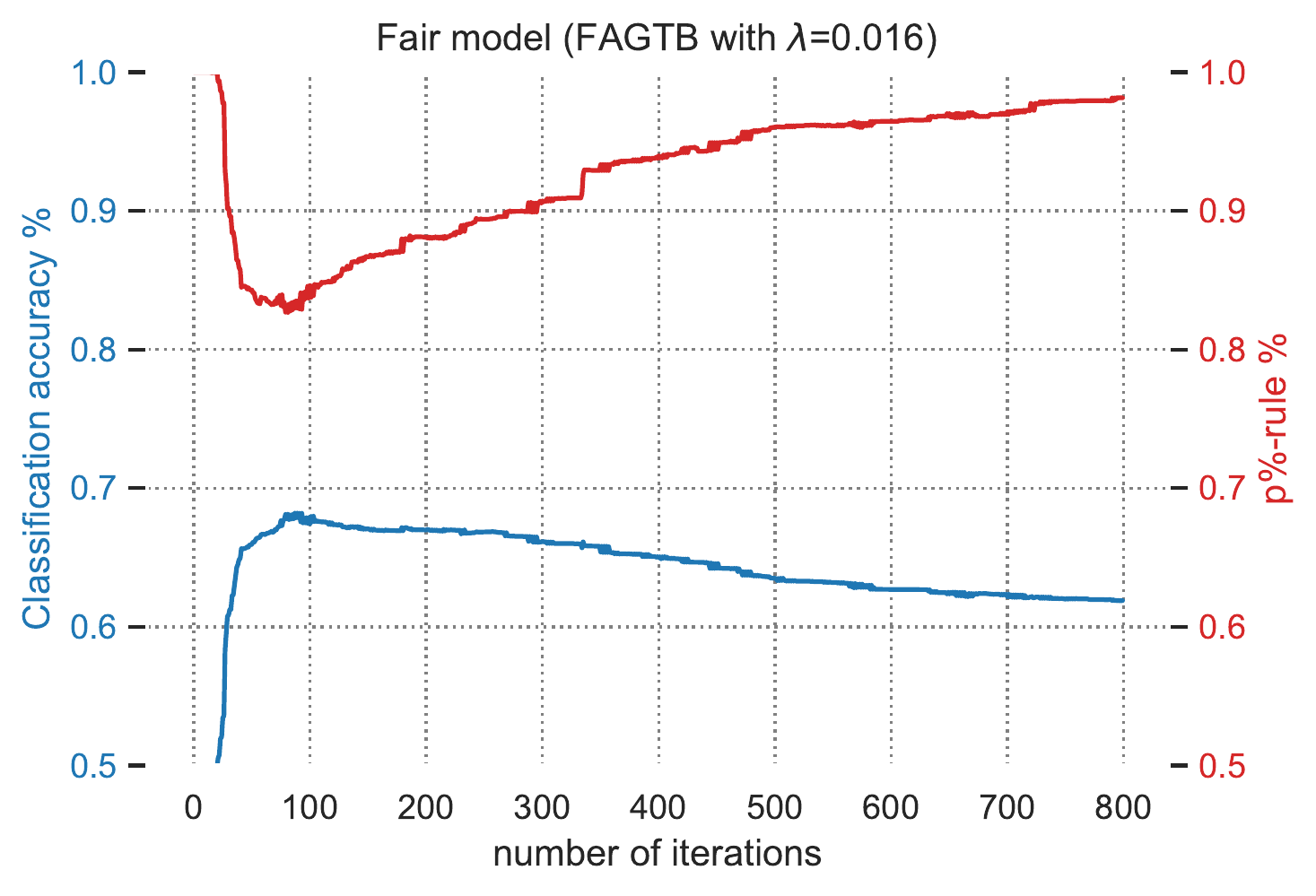}
  \caption{Synthetic scenario: Accuracy and p-rule metric for a biased model ($\lambda=0$) and for several fair models with varying values of $\lambda$ optimized for demographic parity.}
  \label{fig:toy_scenario_accuracy}
\end{figure*}

\section{Empirical Results}
\label{sec:empirical_results}
We evaluate the performance of our algorithm empirically with respect to regression accuracy and fairness. We conduct the experiments on a synthetic scenario, but also on real-world data sets. Finally, we compare the results with state-of-the-art algorithms.

\subsection{Synthetic Scenario}

We illustrate the fundamental functionality of our proposal with a simple toy scenario which was inspired by the Red Car example~\citep{Kusner2017}. The subject is a pricing algorithm for a fictional car insurance policy. The purpose of this exercise is to train a fair classifier which estimates the claim likelihood without incorporating any gender bias. We want to demonstrate the effects of an unfair model versus a fair model.

We focus on the general claim likelihood and ignore the severity or cost of the claim. Further, we only consider the binary case of claim or not (as opposed to a frequency). We assume that the claim likelihood only depends on the aggressiveness and the inattention of the policyholder. To make the training more complex, these two properties are not directly represented in the input data but only indirectly available through correlations with other input features. We create a binary label $Y$ with no dependence with the sensitive attribute $S$. Concretely, we use as features the protected attribute \textit{gender} of the policyholder, and the unprotected attributes \textit{color} of the car, and \textit{age} of the policyholder. In our data distribution, the \textit{color} of the car is strongly correlated with both \textit{gender} and aggressiveness. The \textit{age} is not correlated with \textit{gender}. However, the \textit{age} is correlated with the inattention of the policyholder. Thus, the latter input feature is actually linked to the claim likelihood. 
First, we generate the training samples ${(x_{i},s_{i},y_{i})}_{i=1}^{n}$. The unprotected attributes $x_{i}=(c_{i},a_{i})$ represent the \textit{color} of the car and the \textit{age} of the policyholder, respectively. $s$ is the protected variable \textit{gender}. $y$ is the binary class label, where $y=1$ indicates a registered claim. As stated above, we do not use the two features aggressiveness (A) and inattention (I) as input features but only to construct the data distribution which reflects the claim likelihood. In order to make it more complex, we add a little noise $\epsilon_{i}$. These training samples are generated as follows: For each $i$, let $s$ be a discrete variable with the discrete uniform distribution such that $s_{i} \in [0, 1]$. 

\begin{align*}
\begin{pmatrix} I_{i}\\
a_{i}
\end{pmatrix} &\sim \mathcal{N}
\begin{bmatrix}
\begin{pmatrix}
0\\
40
\end{pmatrix}\!\!,&
\begin{pmatrix}
1 & 4 \\
4 & 20
\end{pmatrix}
\end{bmatrix}\\[2\jot]
A_i &\sim \mathcal{N}(0,1)\\
c_{i} &=  (1.5*s_{i}+A_{i})>1\\
y_{i} &=  \sigma(A_{i}+I_{i}+\epsilon_{i})>0.5\\
\epsilon_{i} &\sim \mathcal{N}(0,0.1)
\end{align*}

A correlation matrix of the distribution is shown in Table~\ref{tab:correlation_matrix}.

We execute first a classical GTB algorithm. In Figure~\ref{fig:toy_scenario_accuracy}, first graph,  we can see the curves of accuracy and the fairness metric p-rule during the training phase. The model shows a stability of the two objectives, this being due to the lack of information and the small number of explanatory variables. Even though there is no obvious link with the sensitive attribute, we notice that this model is unfair (p-rule of 67\%). In fact, the outcome observations $Y$ depend exclusively on $A$ and $I$ which should have no dependence with the sensitive feature $S$. To reconstruct the aggressiveness, the classifier has to consider the color of the car. Unfortunately, it incorporates the sensitive information too, resulting in a claim likelihood prediction one and a half times more for men than for women ($1/0.67$).

To solve this problem and, thus, to achieve demographic parity we use the FAGTB algorithm with a specific hyperparameter $\lambda$. This hyperparameter is obtained by 10-fold cross validation on 20\% of the test set. As explained above, the choice of this value depends on the main objective, resulting in a trade-off between accuracy and fairness. We decided to train a model that reaches a p-rule of approximately 95\% with a $\lambda$ equal to 0.015.  

\begin{figure}[h]
  \centering
  \includegraphics[scale=0.30,valign=t]{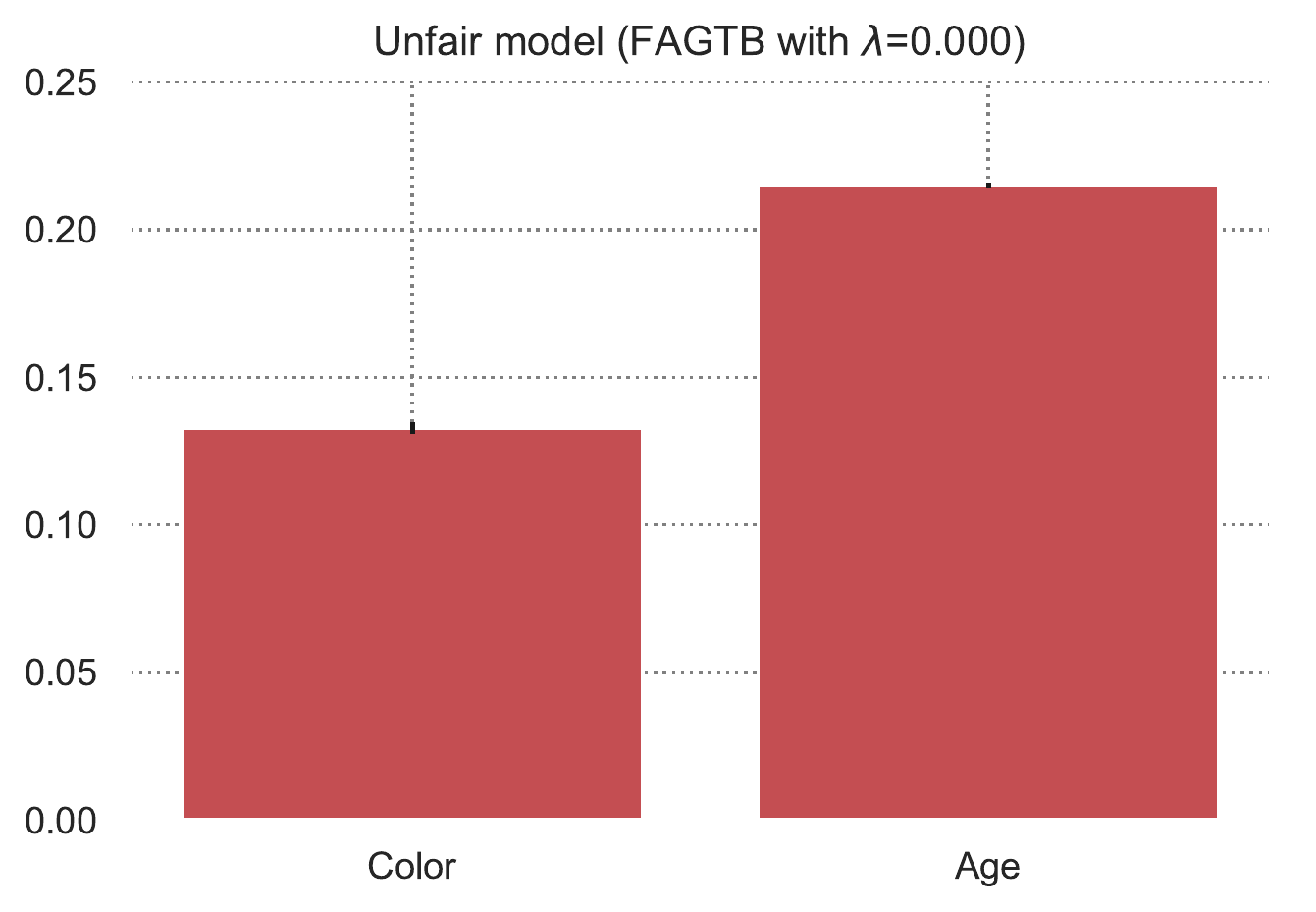}
  \includegraphics[scale=0.30,valign=t]{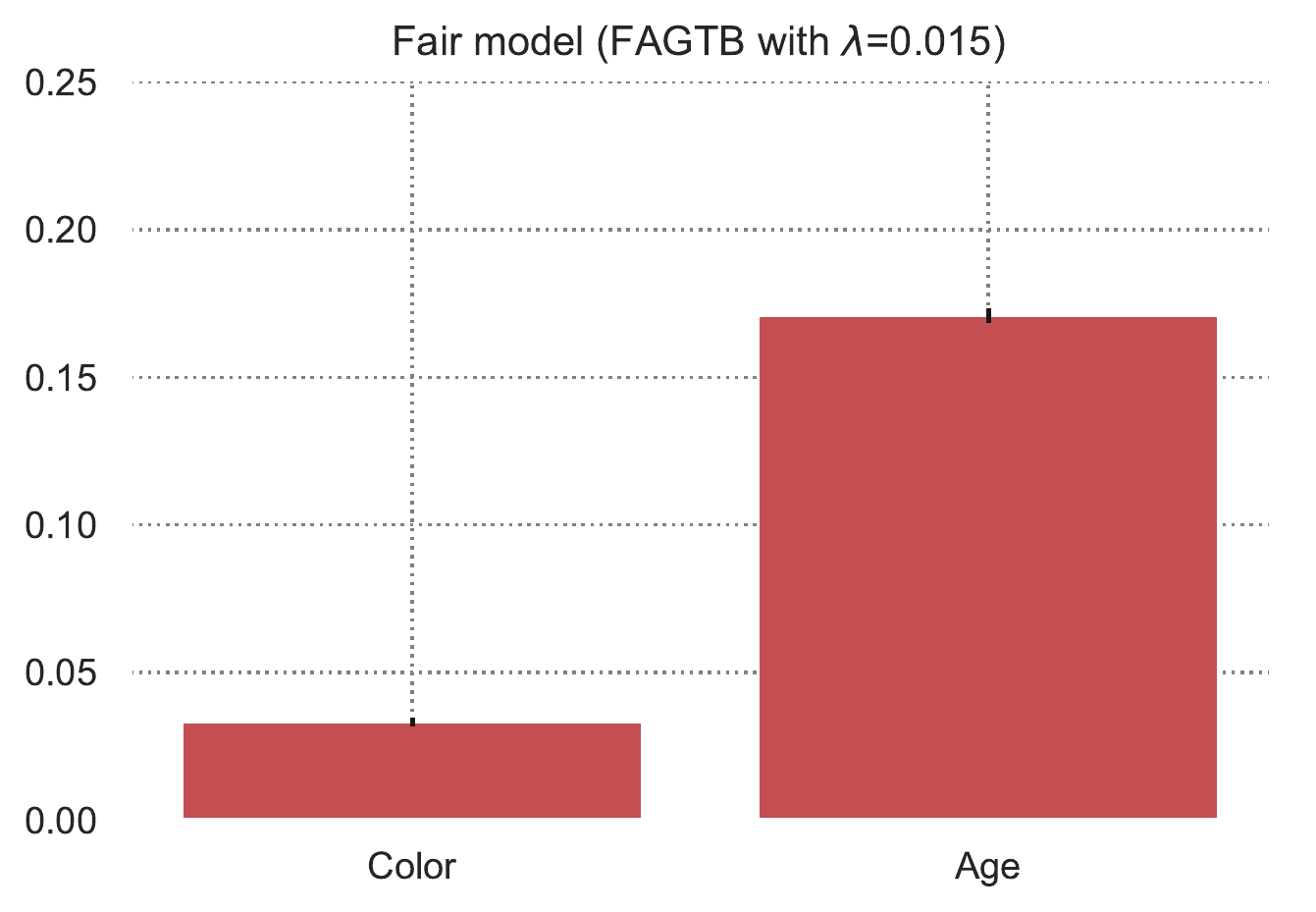}
  \caption{Synthetic Scenario: Feature importance for a biased model ($\lambda=0$) and a fair model ($\lambda=0.015$) optimized for demographic parity.
   \label{fig:toy_scenario_feature_importance}}
\end{figure}

In Figure~\ref{fig:toy_scenario_accuracy}, we also plot 5 others models with different values of $\lambda$, optimized for demographic parity. We observe that during training, when the attenuation of the bias is sudden, the accuracy also dramatically drops. We note that gaining 29 points of p-rule leads to a decrease of accuracy of 10 points. To have a better understanding of what is happening when we consider the model as fair in this specific scenario, we plot the features importance permutation for the fair and the unfair model in Figure~\ref{fig:toy_scenario_feature_importance}.
The model reported importance on the age feature, which is not correlated with the sensitive. 
The difference between the two features is higher for the fair model (0.145 points), the  color feature becoming insignificant. With higher lambda values, the weight of this indirectly correlated feature would tend to 0. 

\begin{table}[h]
\caption{Correlation matrix of the synthetic scenario}
\label{tab:correlation_matrix}

\centering
\begin{tabular}{*{6}{p{1cm}}}

\cline{2-2}
\multicolumn{1}{c|}{a}         & \multicolumn{1}{c|}{1.0}   &                            &                           &                          &                          \\ \cline{2-3}
\multicolumn{1}{c|}{A} & \multicolumn{1}{c|}{0.01}  & \multicolumn{1}{c|}{1.0}   &                           &                          &                          \\ \cline{2-4}
\multicolumn{1}{c|}{c}       & \multicolumn{1}{c|}{-0.01} & \multicolumn{1}{c|}{0.68}  & \multicolumn{1}{c|}{1.0}  &                          & \multicolumn{1}{l}{}     \\ \cline{2-5}
\multicolumn{1}{c|}{s}      & \multicolumn{1}{c|}{0.0}   & \multicolumn{1}{c|}{-0.01} & \multicolumn{1}{c|}{0.36} & \multicolumn{1}{c|}{1.0} & \multicolumn{1}{l}{}     \\ \cline{2-6} 
\multicolumn{1}{c|}{I} & \multicolumn{1}{c|}{0.90}  & \multicolumn{1}{c|}{0.01}  & \multicolumn{1}{c|}{0.0}  & \multicolumn{1}{c|}{0.0} & \multicolumn{1}{c|}{1.0} \\ \cline{2-6} 
\multicolumn{1}{c}{} & \centering{a} & \centering{A} & \centering{c} & \centering{s} & \centering{I}
\end{tabular}
\\[5pt]
The features are: age (a), agressivity (A), color (c), gender (s), inattention (I)
\end{table}

\subsection{Comparison against the State-of-the-Art}
\subsubsection{Data sets}
\label{sec:data sets and fairness algorithms comparisons}

\begin{table*}[h]
\caption{Data sets used for the experiments}
\label{tab:description_of_data sets}
\centering
\begin{tabular}{lllllll}
\hline
\multicolumn{1}{c}{Data set} & \multicolumn{1}{c}{\# Observations} & \multicolumn{1}{c}{\# Features} & \multicolumn{1}{c}{Target} & \multicolumn{1}{c}{\%Target} & \multicolumn{1}{c}{Sensitive} & \multicolumn{1}{c}{\%Sensitive} \\ \hline
Adult UCI                      &                    45,000                 &                   14              &                    Income $>=$ \$50k        &                    30.0\%       &                  Gender         &           58.0\%                           \\
         \\ \hline
COMPAS                       &                    6,967                 &                   13              &                    2-year recidivism  &                    45.5\%       &                  Race         &           34.0\%                          \\
                             \\ \hline
Default                       &                    30,000                 &                   23              &                    Defaulting on payments        &            22.1\%              &                  Gender         &           60.4\%                             \\
                            \\ \hline
Bank                       &                    45,211                 &                   16              &                    Subscription to a term deposit      &                 $11.7\%$      &                  Age         &         $32.9\%$                              \\
                             \\ \hline

\end{tabular}
\\[5pt] 
Description of the data sets: number of observations, number of features, target, total share of the target, sensitive attribute,\\ and total share of the sensitive attribute.
\end{table*}

For our experiments we use 4 different popular data sets often used in fair classification (Table~\ref{tab:description_of_data sets}): 
\begin{itemize}
\item Adult: The Adult UCI income data set~\citep{Dua:2019} contains 14 demographic attributes of approximately 45,000 individuals together with class labels which state if their income is higher than \$50,000 or not. As sensitive attribute we use gender encoded as a binary attribute, male or female.
\item Compas: The COMPAS data set~\citep{angwin2016machine} contains 13 attributes of about 7,000 convicted criminals with class labels that state whether or not the individual recidivated within 2 years of its most recent crime. Here, we use race as sensitive attribute, encoded as a binary attribute, Caucasian or not-Caucasian.
\item Default: The Default data set~\citep{Yeh:2009:CDM:1464526.1465163} contains 23 features about 30,000 Taiwanese credit card users with class labels which state whether an individual will default on payments. As sensitive attribute we use gender encoded as a binary attribute, male or female.
\item Bank: The bank marketing data set~\citep{Moro2014} contains 16 features about 45,211 clients of a Portuguese banking institution. The goal is to predict if the client has subscribed or not to a term deposit. We consider the age as sensitive attribute, encoded as a binary attribute, indicating whether the client's age is between 33 and 60 years, or not.
\end{itemize}



For all data sets, we repeat 10 experiments by randomly sampling two subsets, 80\% for the training set and 20\% for the test set. Finally, we report the average of the accuracy and the fairness metrics from the test set. 

\subsubsection{Fairness algorithms}

Because different optimization objectives result in different algorithms, we run separate experiments for the two fairness metrics of our interest, demographic parity (Table~\ref{tab:results_demographic_parity}) and equalized odds (Table~\ref{tab:results_equalized_odds}). More specifically, for demographic parity we aim at a p-rule of 90\% for all algorithms and then compare the accuracy. Optimizing for equalized odds, results are more difficult to compare. In order to be able to compare the accuracy, we have done our best to obtain, each time, a disparate level below 0.03.  

As a baseline, we use a classical, "unfair" gradient tree boosting algorithm, Standard GTB, and a deep neural network, Standard NN. 

Further, to evaluate if the complexity of the adversarial network has an impact on the quality of the results, we compare a simple logistic regression adversarial, FAGTB-1-Unit, with a complex deep neural network, FAGTB-NN. 

In addition to the algorithms mentioned above, we evaluate the following fair state-of-the-art in-processing algorithms:
Wadsworth2018~\citep{Wadsworth2018}\footnotemark[2],
Zhang2018~\citep{Zhang2018}\footnotemark[3], Kamishima~\citep{Kamishima2012}\footnotemark[1] Feldman~\citep{Feldman2014}\footnotemark[1],
Zafar-DI~\citep{zafar2017parity}\footnotemark[1]
and Zafar-DM~\citep{Zafar2017}\footnotemark[1]. 

\footnotetext[1]{\url{https://github.com/algofairness/fairness-comparison}}
\footnotetext[2]{\url{https://github.com/equialgo/fairness-in-ml}}
\footnotetext[3]{\url{https://github.com/IBM/AIF360}}

\begin{figure}[h]
  \centering
  \includegraphics[scale=0.62,valign=t]{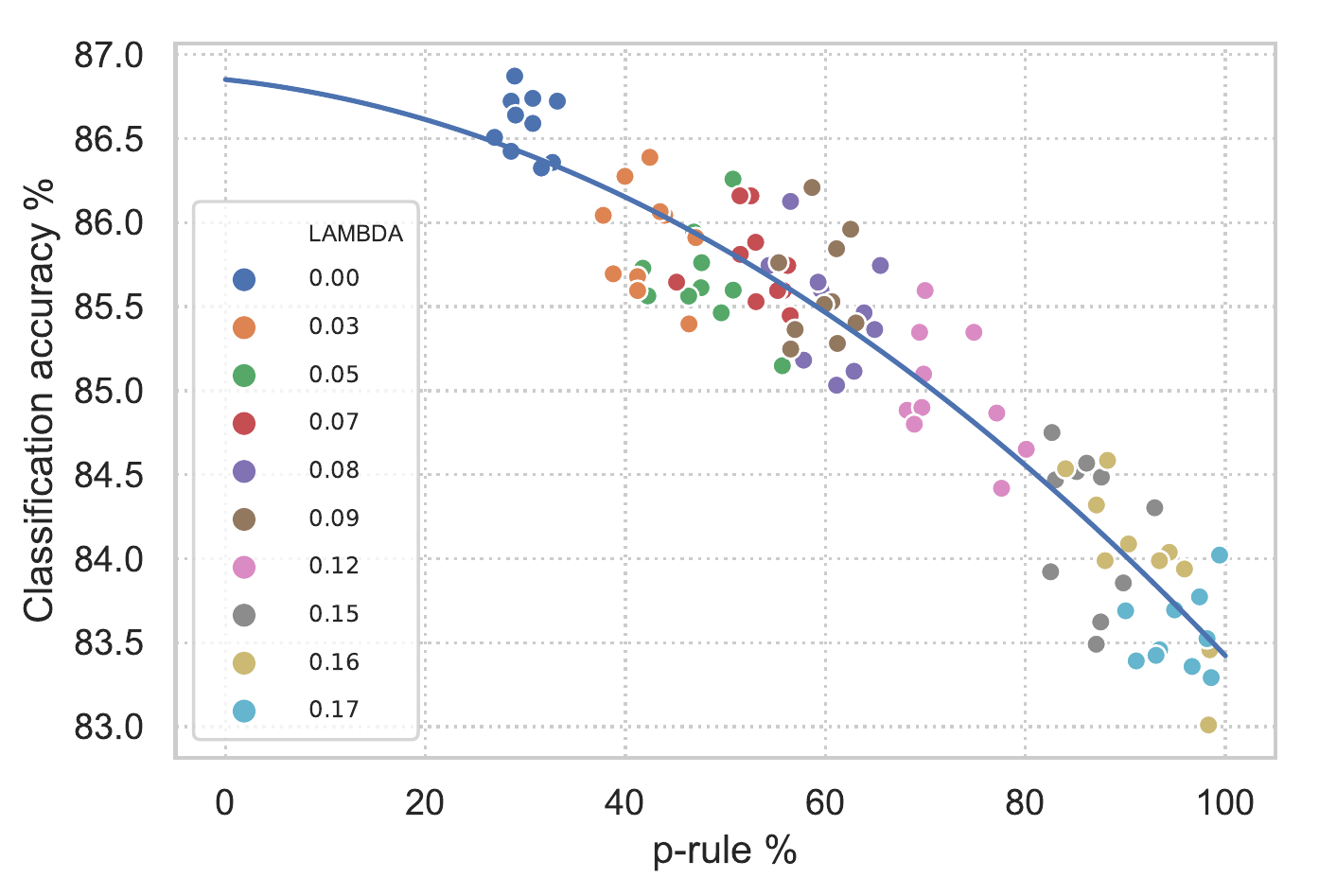}
  \caption{Impact of hyperparameter $\lambda$ (Adult UCI data set): Higher values of $\lambda$ produce fairer predictions, while $\lambda$ near 0 allows to only focus on optimizing the classifier predictor.}
  \label{fig:lambdapruleacc}
\end{figure}

For each algorithm and for each data set, we obtain the best hyperparameters by grid search in 5-fold cross validation (specific to each of them). As a reminder, for FAGTB the $\lambda$ value is used to balance the 2 cost functions during the training phase. This value depends exclusively on the main objective: For example, to obtain the demographic parity objective with 90\% p-rule, we choose a lower and thus less weighty $\lambda$ than for a 100\% p-rule objective. In order to better understand this hyperparameter $\lambda$ we illustrate its impact on the accuracy and the p-rule metric in Figure~\ref{fig:lambdapruleacc} for the Adult UCI data set. For that, we model the FAGTB-NN algorithm with 10 different values of $\lambda$ and we run each experiment 10 times. In the graph, we report the accuracy and the p-rule fairness metric, and finally plot a polynomial regression of second order to demonstrate the general effect.

For Standard GTB, we parameterize the number of trees and the maximum tree depth. For example, for the Bank data set, a tree depth of $3$ with $800$ trees is sufficient.
For the Standard NN, we parameterize the number of hidden layers and units with a ReLU function and we apply a specific dropout regularization to avoid overfitting. Further, we use an Adam optimisation with a binary cross entropy loss. For the Adult UCI data set for example, the architecture consists of 2 hidden layers with 16 and 8 units, respectively, and ReLU activations. The output layer comprises one single output node with sigmoid activation. 

\begin{figure*}[t]
 \centering
 \includegraphics[scale=0.42,valign=t]{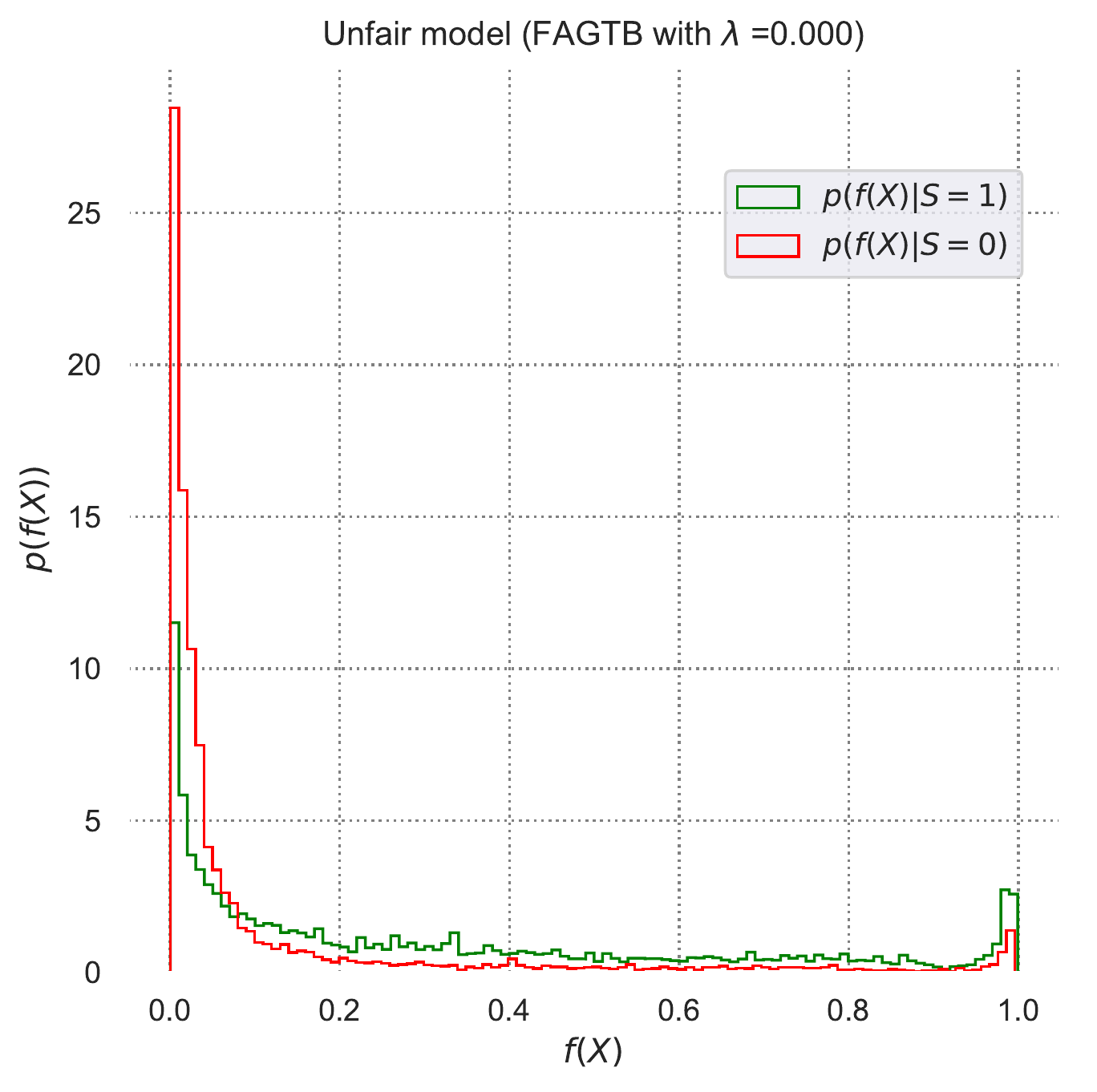}
   \includegraphics[scale=0.42,valign=t]{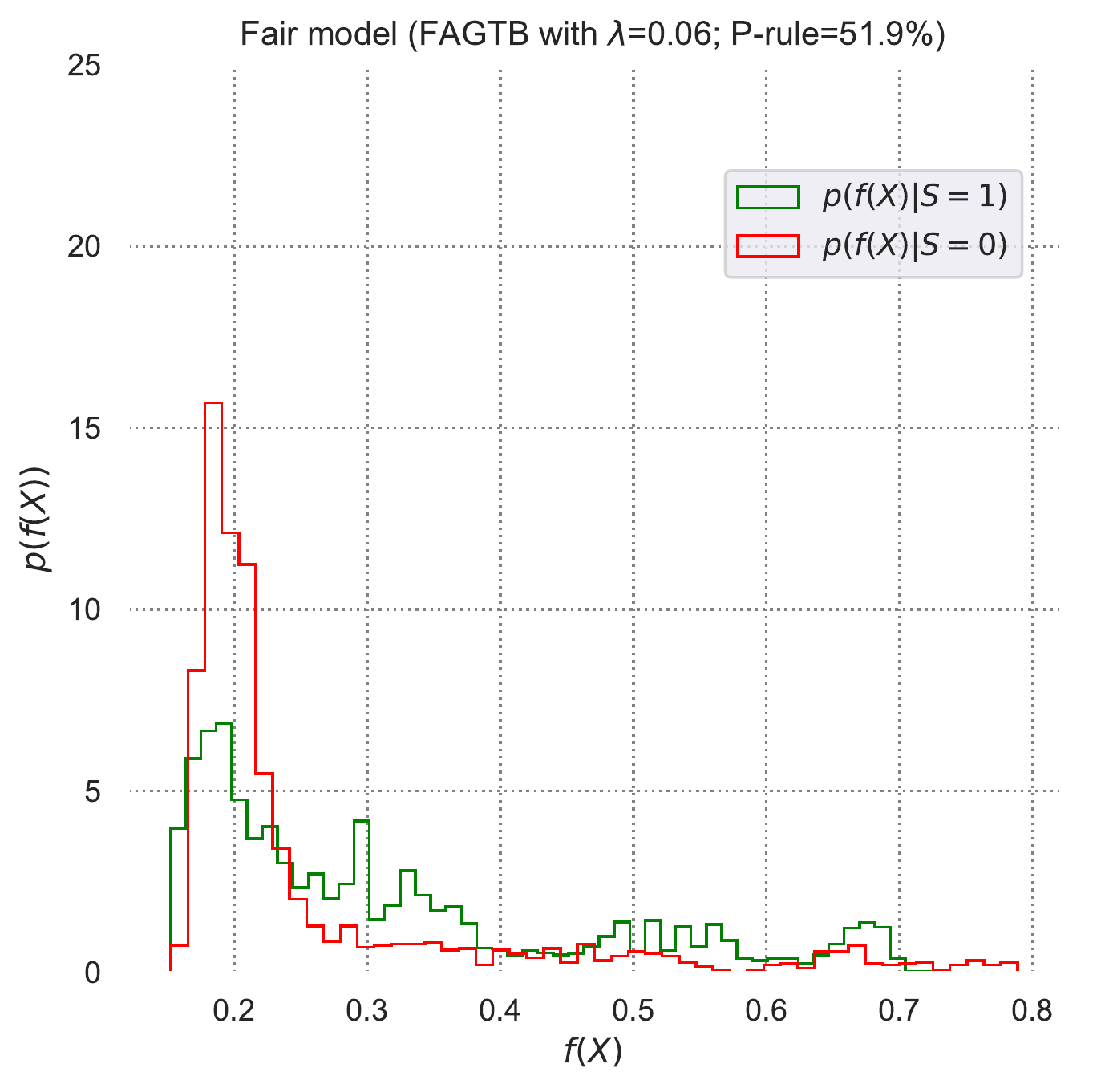}
 \includegraphics[scale=0.42,valign=t]{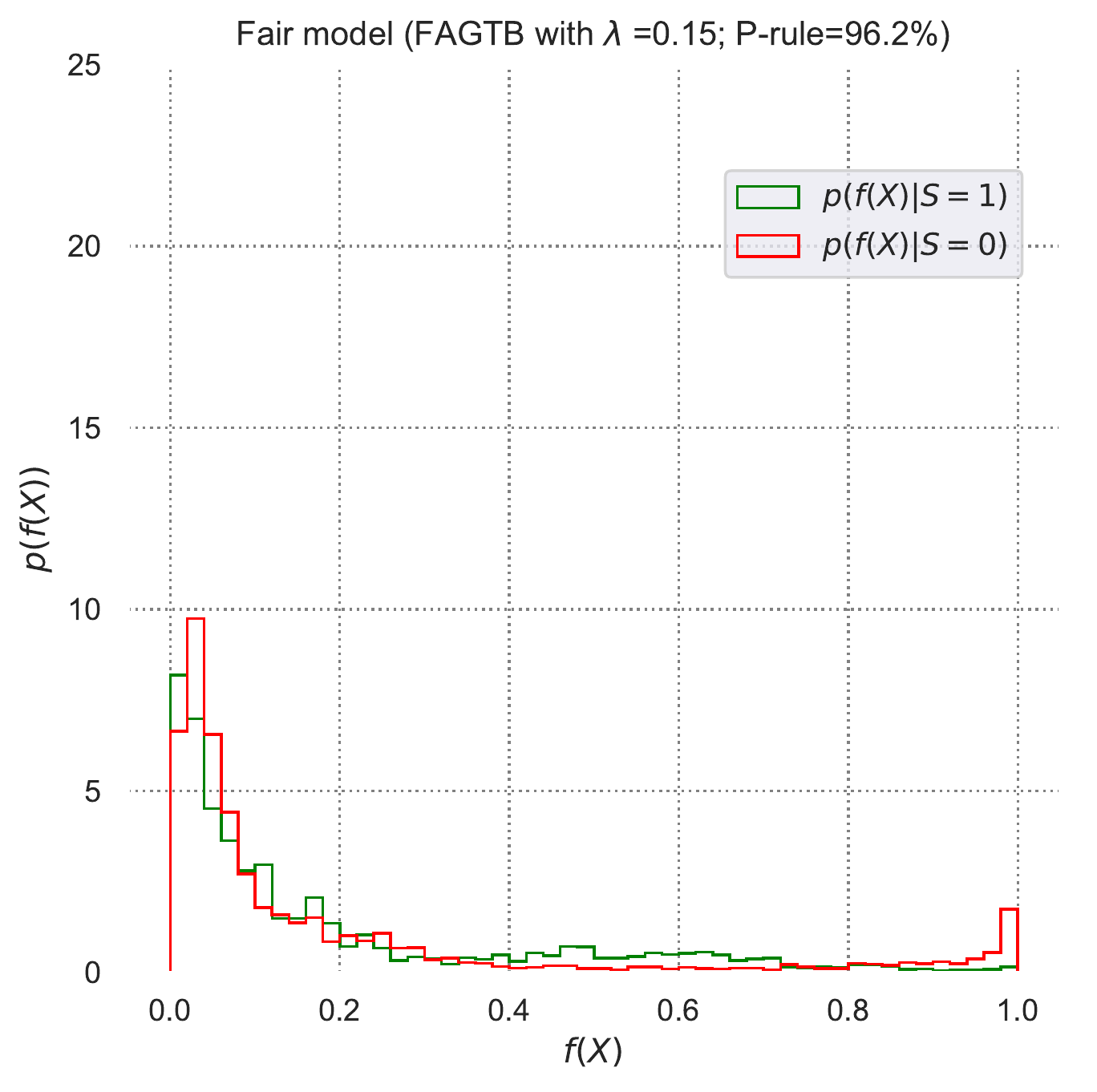}
 \caption{Distributions of the predicted probabilities given the sensitive attribute $S$ (Adult UCI data set)}
 \label{fig:distribution_of_predictions_by_sensitive}
\end{figure*}

For FAGTB, to accelerate the learning phase, we decided to sacrifice some performance by replacing the one-dimensional optimization $\gamma_{m}$ by a specific fixed learning rate for the classifier predictor. All hyperparameters mentioned above, for trees and neural networks, are selected jointly. Notice that those choices impact the rapidity of convergence for each of them. For example, if the classifier predictor converges too quickly this may result in biased prediction probabilities during the first iterations which are difficult to correct by the adversary afterwards. For FAGTB-NN, in order to achieve better results, we execute for each gradient boosting iteration several training iterations of the adversarial NN. This produces a more persistent adversarial algorithm. Otherwise, the predictor classifier GTB could dominate the adversary. At the first iteration, we begin with modeling a biased GTB and we then model the adversarial NN based on those biased predictions. This approach allows to have a better weight initialization of the adversarial NN. It is more suitable for the specific bias on the data set. Without this specific initialization we encountered some cases where the predictor classifier surpasses the adversarial too quickly and tends to dominate from the beginning. Compared to the FAGTB-NN, the adversary of the FAGTB-1-Unit is more simple. In this case, the 2 parameters of the adversarial are chosen randomly and for each gradient boosting iteration only one is computed for the adversarial unit.





\subsubsection{Results}

For demographic parity (Table~\ref{tab:results_demographic_parity}), as expected Standard GTB and Standard NN achieve the highest accuracy. However, they are also the most biased ones. For example, the classical gradient tree boosting algorithm achieves a 32.6\% p-rule for the Adult UCI data set. In this particular case, the prediction for earning a salary above \$50,000 is in average more than three times higher for men than for women. 
Comparing the mitigation algorithms, FAGTB-NN achieves the best result with the highest accuracy while maintaining a reasonable high p-rule equality (90\%). 
The choice of a neural network architecture for the adversary proved to be in any case better than a simple logistic regression. This is particularly true for the COMPAS data set where, for a similar p-rule, the difference in accuracy is considerable (2.7 points). Recall that for demographic parity the adversarial classifier only has one single input feature which is the output of the prediction classifier. It seems necessary to be able to segment this input in several ways to better capture information relevant to predict the sensitive attribute. The sacrifice of accuracy is less important for the Bank and the Default data set. The dependence between the sensitive attribute and the target label is thus less important than for the COMPAS data set. To achieve a p-rule of 90\%, we sacrifice 4.6 points of accuracy (Comparing GTB and FAGTB-NN) for COMPAS, 0.7 points for Default and 0.6 points for Bank.

In Figure~\ref{fig:distribution_of_predictions_by_sensitive} we plot the distribution of the predicted probabilities for each sensitive attribute $S$ for 3 different models: An unfair model with $\lambda=0$, and 2 fair FAGTB models with $\lambda=0.06$ and $\lambda=0.15$, respectively. For the unfair model, the distribution differs most for the lower probabilities. The second graph shows an improvement but there remain some differences. For the final one, the distributions are practically aligned.

Zhang2018~\citep{Zhang2018} introduced a projection term which ensures that the predictor never moves in a direction that could help the adversary. While this is an interesting approach, we noticed that this term does not improve the results for demographic parity. In fact, the Wadsworth2018~\citep{Wadsworth2018} algorithm follows the same approach but without projection term and obtains similar results.


\begin{table*}[h!]
\caption{Results for Demographic Parity}
\label{tab:results_demographic_parity}
\centering
\begin{tabular}{l|l|l|l|l|l|l|l|l|}
\cline{2-9}
& \multicolumn{2}{c|}{Adult}                                  & \multicolumn{2}{c|}{COMPAS}                                & 
\multicolumn{2}{c|}{Default}                                &
\multicolumn{2}{c|}{Bank}   \\ \cline{2-9} 
& \multicolumn{1}{c|}{Accuracy} & \multicolumn{1}{c|}{P-rule} & \multicolumn{1}{c|}{Accuracy} & \multicolumn{1}{c|}{P-rule} & \multicolumn{1}{c|}{Accuracy} & \multicolumn{1}{c|}{P-rule} & Accuracy       & P-rule       \\ \hline
\multicolumn{1}{|l|}{Standard GTB}         &    86.8\%     &   32.6\%    &  69.1\%                             &          61.2\%                      &                  82.9\%                  &          77.2\%                      &        90.8\%        &       48.1\%       \\ \hline
\multicolumn{1}{|l|}{Standard NN}              &         85.3\%                      &            31.4\%                 &             67.5\%                  &            71.1\%                 &              82.1\%                 &              63.3\%             &            90.3\%        &     58.6\%              \\ \hline
\multicolumn{1}{|l|}{FAGTB-1-Unit}  &           84.4\%      &      90.4\%               &     61.8\%        &        90.1\%     &         81.5\%                   &          90.1\%         &           90.1\%     &   90.0\%                  \\ \hline
\multicolumn{1}{|l|}{FAGTB-NN}   &      \textbf{84.9\%}   &     90.3\%        &        \textbf{64.5\%}    &     90.0\%       &             \textbf{82.2\%}    &      90.2\%                  &    \textbf{90.2\%}      &  90.0\%            \\ \hline
\multicolumn{1}{|l|}{Wadsworth2018~\citep{Wadsworth2018}} &   83.1\%               &       89.7\%      &                63.9\%     &      90.1\%              &        81.8\%         &   90.0\%         &   \textbf{90.2\%}    &          90.1\%      \\ \hline
\multicolumn{1}{|l|}{Zhang2018~\citep{Zhang2018}} &          83.3\%        &           90.0\%         &   64.1\%                 &        89.8\%   &                    81.4\%            &    90.0\%                         &    90.0\%                 &   90.0\%                      \\ \hline
\multicolumn{1}{|l|}{Zafar-DI~\citep{Zafar2015}} &             82.2\%                  &          89.8\%                   &                 63.9\%              &           89.7\%                  &               80.7\%                &                  89.8\%           &       89.2\%             &    90.1\%                       \\ \hline
\multicolumn{1}{|l|}{Kamishima~\citep{Kamishima2012}} &          82.3\%                       &     89.9\%                         &           63.8\%                     &         90.0\%                    &            81.1\%                     &         90.0\%                    &        89.6\%         &       89.9\%             \\ \hline
\multicolumn{1}{|l|}{Feldman~\citep{Feldman2014}} &         \multicolumn{1}{c|}{-}                        &        \multicolumn{1}{c|}{-}                      &           61.4\%                       &                90.1\%                  &          72.2\%                          &          90.2\%                      &         \multicolumn{1}{c|}{-}       &        \multicolumn{1}{c|}{-}                \\ \hline
\end{tabular}
\\[5pt] 
Comparing our approach with different common fair algorithms by accuracy and fairness (p-rule metric)\\ for the Adult UCI, the COMPAS, the Default and the Bank data set.
\end{table*}

\begin{table*}[h!]
\caption{Results for Equalized Odds}
\label{tab:results_equalized_odds}
\centering
\subfloat{
\centering
\begin{tabular}{l|l|l|l|l|l|l|}
\cline{2-7}
                                         & \multicolumn{3}{c|}{Adult}   & \multicolumn{3}{c|}{COMPAS} \\ \cline{2-7} 
                                         & Accuracy & DispFPR & DispFNR & Accuracy & DispFPR & DispFNR \\ \hline
\multicolumn{1}{|l|}{Standard GTB}         & 86.8\%  & 0.06  & 0.07   &     69.1\%       &    0.12     &   0.20      \\ \hline
\multicolumn{1}{|l|}{Standard NN}              &   85.3\%   &   0.07      &    0.10     &       67.5\%     &    0.09     &    0.15     \\ \hline
\multicolumn{1}{|l|}{FAGTB-1-Unit}              &     86.3\%      &   0.02       &    0.02   &  65.1\%  &    0.03     &    0.12     \\ \hline
\multicolumn{1}{|l|}{FAGTB-NN}              &      \textbf{86.4\%}         &    0.02        &    0.02      &   \textbf{66.2\%}        &   0.01      &    0.02     \\ \hline
\multicolumn{1}{|l|}{Wadsworth2018~\citep{Wadsworth2018}}               &     84.9\%       &    0.02     &       0.03      &      65.4\%     &   0.02      &   0.03         \\ \hline
\multicolumn{1}{|l|}{Zhang2018~\citep{Zhang2018}}              &     84.8\%       &    0.03     &       0.03      &      64.9\%     &   0.03      &   0.02       \\ \hline
\multicolumn{1}{|l|}{Zafar-DM~\citep{Zafar2017}}   &    83.9\%      &   0.03      &    0.09     &    64.3\%      &    0.09     &   0.17     
\\ \hline
\multicolumn{1}{|l|}{Kamishima~\citep{Kamishima2012}}   & 82.6\%         &   0.06      &     0.24    &      63.6\%    &  0.08       &    0.11     
\\ \hline
\multicolumn{1}{|l|}{Feldman~\citep{Feldman2014}}              &    80.6\%      &     0.07    &   0.05      &   61.1\%       &    0.03     &     0.03    \\ \hline

\end{tabular}
}

\vspace*{0.1cm}

\subfloat{
\centering
\begin{tabular}{l|l|l|l|l|l|l|}
\cline{2-7}
                                         & \multicolumn{3}{c|}{Default}   & \multicolumn{3}{c|}{Bank} \\ \cline{2-7} 
                                         & Accuracy & DispFPR & DispFNR & Accuracy & DispFPR & DispFNR \\ \hline
\multicolumn{1}{|l|}{Standard GTB}     &  82.9\%  &  0.02   &  0.04   &   90.8\%   &      0.04    &      0.06      \\ \hline
\multicolumn{1}{|l|}{Standard NN}       &  82.1\%     &  0.02        &  0.05    &     90.3\%     &  0.05    &  0.08    \\ \hline
\multicolumn{1}{|l|}{FAGTB-1-Unit}             &  82.1\%    &  0.00   &   0.01   &   89.7\%   &  0.02  &   0.07   \\ \hline
\multicolumn{1}{|l|}{FAGTB-NN}              &  \textbf{82.5\%}   &  0.00   &  0.01     &  \textbf{90.3\%}  &  0.01  &   0.07  \\ \hline
\multicolumn{1}{|l|}{Wadsworth2018~\citep{Wadsworth2018}}       &  81.2\%   &    0.01 &   0.02     &  89.4\%  &  0.01    &    0.07   \\ \hline
\multicolumn{1}{|l|}{Zhang2018~\citep{Zhang2018}}      &   81.9\%     &    0.00      &    0.01      &    89.8\%    &     0.00     &    0.07       \\ \hline
\multicolumn{1}{|l|}{Zafar-DM~\citep{Zafar2017}}    &  81.0\%   &  0.00   &  0.03    &    89.5\%  &   0.01   & 0.08    
\\ \hline
\multicolumn{1}{|l|}{Kamishima~\citep{Kamishima2012}}    & 80.5\%    &  0.00   &   0.04   &     89.3\%     &  0.00    &   0.08  
\\ \hline
\multicolumn{1}{|l|}{Feldman~\citep{Feldman2014}}      &  71.8\%       &  0.02   &  0.02   &     87.1\%     &     0.05     &   0.06      \\ \hline
\end{tabular}
}
\\
Comparing our approach with different common fair algorithms by accuracy and fairness (DispFPR, DispFNR)\\ for the Adult UCI, the COMPAS, the Default and the Bank data set.
\end{table*}

For equalized odds, the min-max optimization is more difficult to achieve than demographic parity. The fairness metrics DispFPR and DispFNR are not exactly comparable thus we did not succeed to obtain the same level of fairness. However, we notice that the FAGTB-NN achieves better accuracy with a reasonable level of fairness. Concretely, we achieve for the 4 data sets and for both metrics values below 0.02 or less, except for the Bank data set where DispFNR is equal to 0.07. For this data set, most of the state-of-the-art algorithms result in a DispFNR between 0.06 and 0.08. The reason why it proves hard to achieve a low False Negative Rate (FNR), is that the total share of the target is very low at 11.7\%. A possible way to handle this problem of imbalanced target class could be to to add a specific weight directly in the loss function.
We also notice that the difference in the results between FAGTB-1-Unit and FAGTB-NN is much more significant, one possible reason is that an unique logistic regression cannot keep a sufficient amount of information in order to predict the sensitive attribute. 





\section{Conclusion}
\label{sec:conclusion}

In this work, we developed a new approach to produce fair gradient boosting algorithms. Compared with other state-of-the-art algorithms, our method proves to be more efficient in terms of accuracy while obtaining a similar level of fairness. 

Currently, we use a neural network architecture for the adversary. We chose this approach in order to recover the gradient of the input. Another possible strategy is to replace the adversarial neural network by deep neural decision forests~\cite{kontschieder2015deep} which allow to recover the gradient by derivative. Such an architecture, composed of trees only, would make it easier to interpret the model and in particular the role of the sensitive attribute. Another field left for further investigations is the mathematical identification of the optimal hyperparameter $\lambda$. Objectives here are a better convergence of the algorithm and the optimization of the trade-off between accuracy and fairness. 
Finally, it might be interesting to investigate a measure which does not only consider the general case of bias but can also spot and quantify bias that persists on specific sub segments of the population.


\bibliography{MyCollection}


\bibliographystyle{IEEEtran}
\end{document}